# Dynamic Arthroscopic Navigation System for Anterior Cruciate Ligament Reconstruction Based on Multi-level Memory Architecture

**Shuo Wang** [1, †], **Weili Shi** [2, †], **Shuai Yang** [2], **Jiahao Cui** [3] **and Qinwei Guo** [2,*]

1    Department of Engineering Physics, Key Laboratory of Particle and Radiation Imaging, Ministry of Education, Tsinghua University, Beijing 100084, China; shuo-wan19@mails.tsinghua.edu.cn(S.W.)

2    Department of Sports Medicine, Peking University Third Hospital, Institute of Sports Medicine of Peking University, Beijing Key Laboratory of Sports Injuries, Haidian District, Beijing 100191, China; guoqin-wei@vip.sina.com(Q.G.); shiweilixmu@126.com(W.S.); yangshuaibd@pku.edu.cn(S.Y.)

3    State Key Laboratory of Virtual Reality Technology and Systems, Beihang University, Beijing 100191, China; cuijh@buaa.edu.cn(J.C.)

\*    Correspondence: guoqinwei@vip.sina.com (Q.G.)

†    These authors contributed equally to this work.

**Abstract**

This paper presents a dynamic arthroscopic navigation system based on multi-level memory architecture for anterior cruciate ligament (ACL) reconstruction surgery. The system extends our previously proposed markerless navigation method from static image matching to dynamic video sequence tracking. ACL reconstruction surgery typically lasts 60-90 minutes, generating lengthy video sequences that require continuous interpretation and navigation, while the arthroscopic environment presents unique challenges including narrow and constantly changing field of view, specular reflections from moist tissue surfaces, image blurring caused by irrigation fluid, and frequent occlusion of target structures by surgical instruments. By integrating the Atkinson-Shiffrin memory model's three-level memory architecture (sensory memory, working memory, and long-term memory), our system maintains continuous tracking of the femoral condyle throughout the surgical procedure, providing stable navigation support even in complex situations involving viewpoint changes, instrument occlusion, and tissue deformation. Unlike existing methods, our system operates in real-time on standard arthroscopic equipment without requiring additional tracking hardware, achieving 25.3 FPS with a latency of only 39.5 ms, representing a 3.5-fold improvement over our previous static system. For extended sequences (1000 frames), the dynamic system maintained an error of 5.3±1.5 pixels, compared to the static system's 12.6± 3.7 pixels—an improvement of approximately 45%. For medium-length sequences (500 frames) and short sequences (100 frames), the system achieved approximately 35% and 19% accuracy improvements, respectively. Experimental results demonstrate that the system overcomes the limitations of traditional static matching methods, providing new technical support for improving surgical precision and reducing complications in ACL reconstruction.

## 1. Introduction

ACL reconstruction is a common arthroscopic procedure whose success largely depends on precise tunnel placement [1]. Traditional ACL reconstruction surgery relies on the surgeon's experience and anatomical knowledge, lacking objective intraoperative navigation assistance [2]. To address this issue, we previously proposed a markerless navigation system [17] that achieved intraoperative navigation by matching preoperative CT images with intraoperative arthroscopic images. However, this system only supported static image matching and could not adapt to dynamic changes during the surgical procedure.

In actual surgery, arthroscopic imaging serves as the primary visual feedback mechanism, guiding surgeons throughout the entire procedure. ACL reconstruction typically lasts 60-90 minutes, generating lengthy video sequences that require continuous interpretation and navigation. The arthroscopic view constantly changes, requiring real-time adjustment of navigation references. Additionally, the arthroscopic environment presents unique challenges including: narrow and constantly changing field of view, specular reflections from moist tissue surfaces, image blurring caused by irrigation fluid, and frequent occlusion of target structures by surgical instruments. These factors make continuous tracking of anatomical structures in dynamic arthroscopic videos extremely challenging. Existing navigation systems mostly rely on static image matching or external tracking devices, which cannot effectively address these challenges.

Recent years have witnessed remarkable progress in the field of video object segmentation (VOS). Multi-level memory architectures have shown excellent performance in processing long video sequences, maintaining stable tracking despite viewpoint changes and target occlusions. These advances offer new possibilities for solving dynamic tracking problems in arthroscopic navigation, particularly for lengthy procedures like ACL reconstruction where maintaining consistent anatomical reference points is crucial.

This paper presents a novel dynamic arthroscopic navigation system that employs a multi-level memory architecture specifically designed for the arthroscopic environment. Unlike existing methods, our system maintains continuous tracking of the femoral condyle throughout the surgical procedure, providing stable navigation support even in complex situations involving viewpoint changes, instrument occlusion, and tissue deformation. The system operates in real-time on standard arthroscopic equipment without requiring additional tracking hardware, making it practical for clinical implementation.

By addressing the dynamic tracking problem, our system fills an important gap in existing ACL reconstruction navigation technology, providing new technical support for improving surgical precision and reducing complications.

## 2. Related Work

### 2.1 ACL Reconstruction Navigation Systems

Computer-assisted navigation for ACL reconstruction, introduced in 1995, primarily serves to improve bone tunnel positioning accuracy and evaluate post-reconstruction knee joint kinematics. Early prospective randomized controlled studies by Mauch et al. demonstrated that navigation groups achieved tibial tunnel centers closer to ideal positions with significantly reduced graft impingement rates postoperatively [3]. Tensho et al. reported the first case of CT-based computer-navigated ACL reconstruction using skin-based reference points for registration, achieving accuracy within 1mm [4]. However, Meuffels et al. found no significant difference in tunnel placement accuracy between navigation and traditional methods when performed by experienced surgeons, highlighting the importance of surgical expertise [5]. Notably, almost all existing computer-assisted navigation systems rely on marker-based technology, requiring specific markers to be installed on patients and surgical instruments, which inevitably increases surgical complexity and necessitates additional incisions [6, 7].

Reality-based technologies including virtual reality (VR), augmented reality (AR), and mixed reality (MR) represent the newer frontier in ACL reconstruction navigation. VR technology is primarily utilized for assessing ACL injury risk, evaluating postoperative tunnel positions, and designing rehabilitation protocols [8]. AR technology overlays virtual information onto the real environment, showing significant advantages in preoperative planning with the best reported positioning accuracy of 0.32mm, while most other techniques achieve accuracy within 3mm [9]. MR technology offers a novel approach for precise positioning, resulting in more optimal and consistent postoperative tunnel placement [10]. Despite their potential, current reality-based systems still depend on pre-placed markers for accurate registration between virtual and real environments, limiting their application in minimally invasive surgery [11, 12].

A critical limitation shared across nearly all navigation systems is their dependence on marker-based technology, which increases surgical complexity and restricts clinical applications. Future research should focus on developing markerless navigation systems that achieve navigation through direct analysis of intraoperative images, thereby reducing invasiveness and improving user-friendliness [13]. Additionally, long-term follow-up studies are essential to evaluate the clinical efficacy of navigation technologies, as current research indicates that while navigation techniques improve bone tunnel positioning accuracy, they have not yet demonstrated significant differences in postoperative functional scores and clinical outcomes compared to traditional techniques [14, 15]. The integration of artificial intelligence and advanced image processing may further enhance the precision and applicability of navigation systems in ACL reconstruction surgery [16].

Our previous research introduced a marker-less navigation system for anterior cruciate ligament reconstruction that integrated 3D femoral modeling with arthroscopic guidance [17]. This system addressed the critical challenge of accurate femoral tunnel positioning during ACLR procedures by implementing a comprehensive workflow as illustrated in the figure.

The system processed preoperative CT data to generate a detailed 3D model of the distal femur, establishing a standardized Bernard & Hertel (B&H) grid through advanced image processing techniques. A key innovation was the curvature-based feature extraction approach that precisely identified the Capsular Line Reference on the lateral condyle surface. The two-stage registration framework, combining SIFT-ICP algorithms, achieved accurate alignment between the preoperative 3D model and intraoperative arthroscopic views, projecting the B&H grid onto real-time arthroscopic images to provide visual guidance for tunnel placement.

Unlike existing navigation solutions, this system maintained standard surgical workflow without requiring additional surgical instruments or markers. Validation results demonstrated high precision with mean deviation distances of 1.12-1.86 mm. However, this initial system only supported static image matching, requiring recalibration for each new view and limiting its application in dynamic surgical environments.

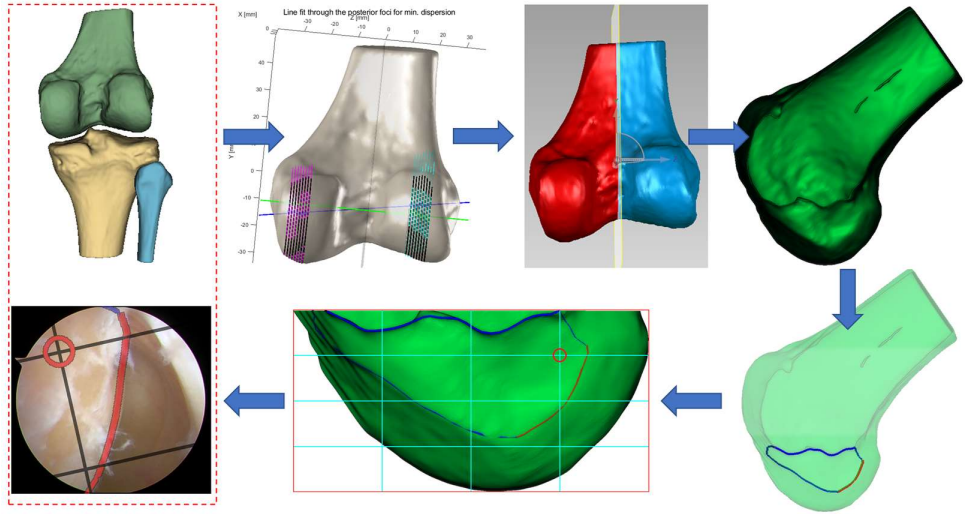

Figure 1. Workflow of the marker-less navigation system for ACL reconstruction. The process begins with a 3D knee model (top left) showing femur (green), tibia (yellow), and fibula (blue). A line fit through posterior foci for minimum dispersion is applied (top center), followed by condyle segmentation into medial (red) and lateral (blue) components (top right). The resulting 3D femoral model (green) undergoes further processing to establish reference grids (bottom right and center), which are ultimately registered with intraoperative arthroscopic images (bottom left) to provide precise guidance for tunnel placement during ACL reconstruction surgery.

**2.2 Video Object Segmentation and Tracking**

SAM [18] has made remarkable advancements in the field of Visual Object Tracking (VOT). Building upon this foundation, the Track Anything Model (TAM) [19] emerged as an innovative solution that delivers exceptional interactive tracking and segmentation capabilities in video sequences. This research introduces Track-Anything, a sophisticated toolkit designed for high-performance object tracking and video segmentation that distinguishes itself through its interactive initialization approach and integration of both SAM and XMem [20]. TAM's architectural framework demonstrates how it effectively leverages SAM's powerful segmentation capabilities as a large-scale model alongside XMem's advanced semi-supervised VOS functionality.

Despite its impressive performance in general video scenarios, the TAM model faces significant limitations when applied to arthroscopic imaging. Arthroscopic videos present unique challenges including highly variable lighting conditions, fluid interference, limited field of view, and rapid camera movements that create motion blur. TAM was primarily trained on natural scene videos with consistent lighting and clear object boundaries, making it poorly suited for the low-contrast, specular reflection-heavy environment of arthroscopic procedures.

Additionally, arthroscopic procedures require real-time processing with minimal latency, while TAM's computational demands from combining both SAM and XMem create processing delays unacceptable in surgical settings. The anatomical structures in arthroscopic views also differ substantially from TAM's training data, with tissue deformation and instrument interactions creating tracking scenarios the model wasn't designed to handle. Furthermore, the interactive initialization approach of TAM is impractical during surgical procedures where surgeon attention must remain focused on the operation rather than model guidance.

For long video processing, Cheng et al. [20] proposed the XMem model based on the Atkinson-Shiffrin memory model, introducing a three-level memory architecture: sensory memory, working memory, and long-term memory. This architecture effectively addresses the balance between memory consumption and accuracy in long video processing, providing theoretical support for our dynamic arthroscopic navigation system.

The methodology presented demonstrates exceptional robustness and user accessibility even in challenging scenarios, offering versatile applications across numerous domains. The research also provides a thorough examination of failure cases, suggesting optimal remediation strategies. With its efficient approach to complex video object perception challenges, this integrated system represents a significant advancement in the field, combining the strengths of both SAM's segmentation precision and XMem's temporal consistency capabilities.

These domain-specific challenges necessitate specialized solutions tailored specifically for arthroscopic imaging rather than general-purpose tracking models like TAM. Our approach adapts key concepts from these advanced models while addressing the unique requirements of surgical navigation environments.

## 3. Methods

### 3.1 System Overview

Our arthroscopic navigation system for ACL reconstruction provides real-time tracking and augmented reality guidance by integrating computer vision algorithms with anatomical knowledge. As illustrated in Figure 2, the system architecture consists of four key components: automatic foreground segmentation, multi-level memory tracking, virtual camera parametrization, and B&H grid projection.

The workflow begins with automatic foreground segmentation (Section 3.2), which employs a hierarchical framework combining HSV color space thresholding with deep learning-based refinement to accurately extract the femoral condyle region from the initial arthroscopic frame. This segmentation serves as the foundation for subsequent tracking and navigation.

The virtual camera parametrization component (Section 3.3) precisely models the optical characteristics of the arthroscope, including intrinsic parameters, distortion coefficients, view frustum, and angular offset calibration. Our feature-priority registration method ensures accurate alignment between arthroscopic images and virtual renderings by prioritizing the matching of the articular margin boundary while incorporating spatial prior constraints.

The core tracking mechanism utilizes a novel multi-level memory architecture (Section 3.4) specifically designed for the challenging arthroscopic environment. This architecture comprises three interconnected memory stores: arthroscopic sensory memory for short-term temporal continuity, arthroscopic working memory for medium-term feature storage with viewpoint information, and arthroscopic long-term memory for persistent feature storage with anatomical identifiers. This design ensures robust tracking performance despite the unique challenges of arthroscopic surgery, including narrow field of view, specular reflections, image blurring from irrigation fluid, and frequent instrument occlusion.

Finally, the B&H grid projection component (Section 3.5) enables precise mapping of standardized anatomical grids from the sagittal plane to the actual arthroscopic view. This transformation facilitates accurate tunnel placement guidance by providing surgeons with familiar anatomical references directly in the arthroscopic field of view.

The integration of these components creates a comprehensive navigation system that maintains continuous tracking of the femoral condyle throughout the surgical procedure while providing clinically relevant augmented reality guidance. The system operates in real-time on standard arthroscopic equipment without requiring additional tracking hardware, making it practical for clinical implementation.

As shown in Figure 2, the integration of these components creates a comprehensive navigation system that maintains continuous tracking of the femoral condyle throughout the surgical procedure while providing clinically relevant augmented reality guidance. The system operates in real-time on standard arthroscopic equipment without requiring additional tracking hardware, making it practical for clinical implementation.

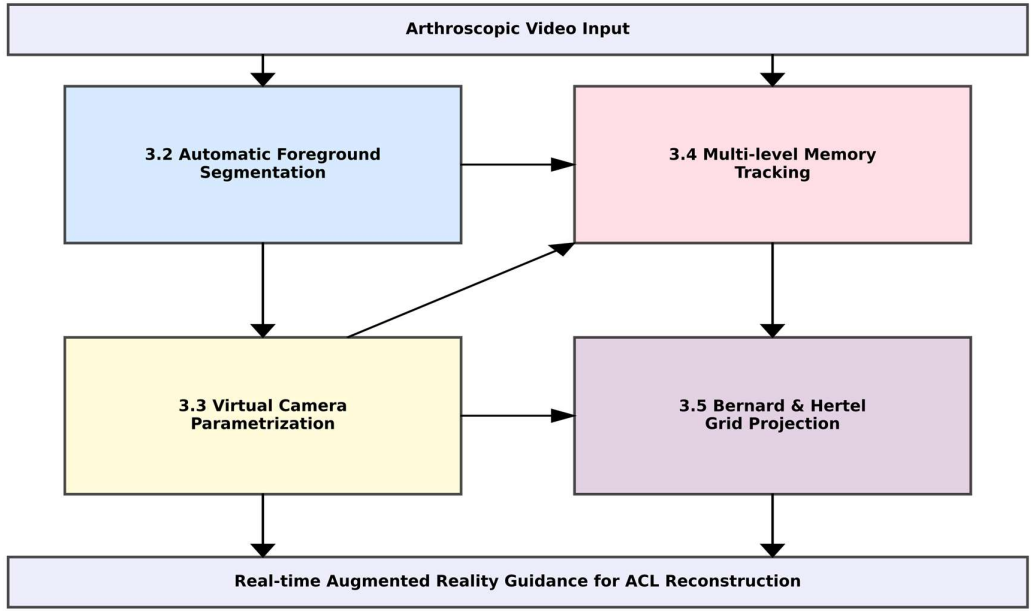

Figure 2. System Architecture for Arthroscopic Navigation in ACL Reconstruction.

### 3.2 Automatic Foreground Segmentation

To initialize the tracking process, we first perform automatic foreground segmentation on the first frame of the arthroscopic video sequence. We propose a hierarchical segmentation framework that integrates traditional image processing techniques with deep learning-based semantic segmentation to achieve robust femoral condyle extraction.

#### 3.2.1 Hierarchical Segmentation Framework

Our segmentation pipeline consists of a multi-stage process that can be formalized as follows:

$$S(I) = \mathcal{F}_{SAM}\left(P(I), C\left(M_{HSV}(I)\right)\right) \tag{1}$$

where $S(I)$ represents the final segmentation result for input image $I$, $\mathcal{F}_{SAM}$ denotes the Semantic-SAM model, $P$ is the image preprocessing function, $C$ extracts the centroid coordinates, and $M_{HSV}$ performs initial mask generation through HSV color space thresholding.

### 3.2.2 Initial Foreground Extraction

We first transform the arthroscopic image $I$ into HSV color space to better separate tissue structures based on their chromatic and illumination properties:

$$I_{HSV} = T_{HSV}(I) \tag{2}$$

where $T_{HSV}$ represents the non-linear transformation from RGB to HSV color space, which helps better isolate the chromatic properties of tissue from illumination conditions in arthroscopic images.

The cartilage region is then extracted using a dual-threshold operation in the HSV space:

$$M_{cart} = \{p \in I_{HSV} | L_H \leq H(p) \leq U_H \wedge L_S \leq S(p) \leq U_S \wedge L_V \leq V(p) \leq U_V\} \tag{3}$$

where $p$ represents a pixel in the HSV image $I_{HSV}$, $H(p)$, $S(p)$, and $V(p)$ denote the hue, saturation, and value components of pixel $p$ respectively. $L_H$, $L_S$, $L_V$ and $U_H$, $U_S$, $U_V$ represent the lower and upper bounds for hue, saturation, and value channels respectively. The symbol $\wedge$ denotes the logical AND operation. For femoral condyle extraction, we empirically set these thresholds to capture the characteristic whitish, bright, low-saturation appearance of cartilage tissue.

### 3.2.3 Centroid Computation

From the initial binary mask $M_{cart}$, we extract the largest connected component $\Omega$ using contour analysis:

$$\Omega = \arg\max_{C_i \in C(M_{cart})} Area(C_i) \tag{4}$$

where $C(M_{cart})$ denotes the set of all contours in $M_{cart}$. The centroid coordinates $(c_x, c_y)$ of $\Omega$ are then computed using spatial moments:

$$c_x = \frac{m_{10}}{m_{00}}, c_y = \frac{m_{01}}{m_{00}} \tag{5}$$

where $m_{pq}$ represents the $(p,q)$-order spatial moment of $\Omega$, with $p$ and $q$ being the powers of $x$ and $y$ coordinates respectively:

$$m_{pq} = \sum_{(x,y) \in \Omega} x^p y^q \tag{6}$$

### 3.2.4 Deep Learning-based Refinement

We employ a Semantic-SAM model pre-trained on arthroscopic image datasets to refine the segmentation. The model $\mathcal{F}_{SAM}$ takes the original image $I$ and the normalized centroid coordinates $\left(\frac{c_x}{W}, \frac{c_y}{H}\right)$ as inputs:

$$\{M_1, M_2, \cdots, M_k\} = \mathcal{F}_{SAM}\left(I, \left(\frac{c_x}{W}, \frac{c_y}{H}\right)\right) \tag{7}$$

where $W$ and $H$ are the image width and height, respectively, and $\{M_1, M_2, \cdots, M_k\}$ is the set of candidate segmentation masks.

### 3.2.5 Optimal Mask Selection

The final segmentation mask $M^*$ is selected by maximizing the Intersection over Union ($IoU$) with the initial mask $M_{cart}$:

$$M^* = \arg\max_{M_i \in \{M_1, M_2, \cdots, M_k\}} IoU(M_i, M_{cart}) \tag{8}$$

where the $IoU$ is defined as:

$$IoU(M_i, M_{cart}) = \left|\frac{M_i \cap M_{cart}}{M_i \cup M_{cart}}\right| \tag{9}$$

As shown in Figure 3, the proposed system integrates traditional image processing techniques with deep learning for accurate cartilage segmentation. The process begins with an arthroscopic input image (left), which undergoes HSV color space transformation to enhance feature discrimination. Initial mask generation through HSV thresholding identifies the cartilage region, followed by centroid computation to locate the region of interest. The centroid serves as a prompt for Semantic-SAM segmentation, generating multiple candidate masks. These candidates undergo IoU comparison with the initial mask to select the optimal segmentation result. The final output (bottom left) shows the segmented cartilage region overlaid on the original image, providing clinically relevant visualization for surgical navigation and assessment.

This automatic segmentation method eliminates the need for manual annotation, enhancing the system's automation and clinical usability. The segmentation result serves as the initial state for tracking, providing a foundation for subsequent dynamic tracking. The integration of traditional image processing for initial estimation with deep learning-based refinement ensures both efficiency and accuracy, which are crucial for real-time surgical navigation.

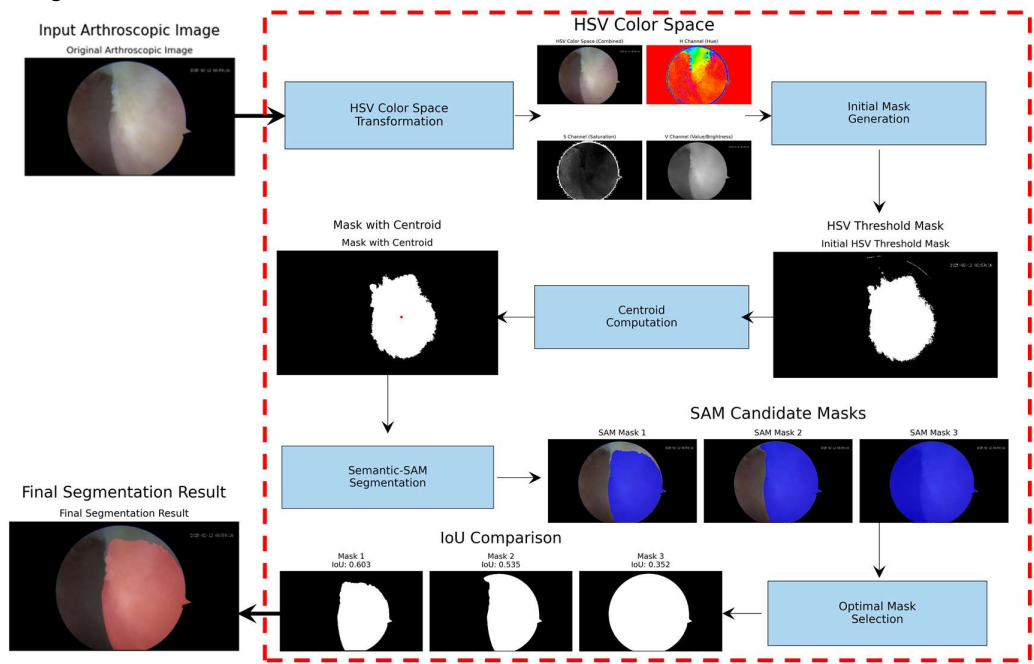

Figure 3. Framework of Arthroscopic Cartilage Segmentation System.

### 3.3 Virtual Camera Parametrization for Arthroscopic Rendering

The accurate segmentation of the femoral condyle established in Section 3.2 provides the foundation for tracking, but achieving realistic augmented reality visualization requires precise alignment between the arthroscopic view and virtual rendering. Section 3.3 addresses this challenge through virtual camera parametrization, as shown in Figure 4, where we model the optical characteristics of the arthroscope—including intrinsic parameters, distortion coefficients, view frustum, and angular offset—to ensure visual consistency between real and virtual elements in the augmented reality environment.

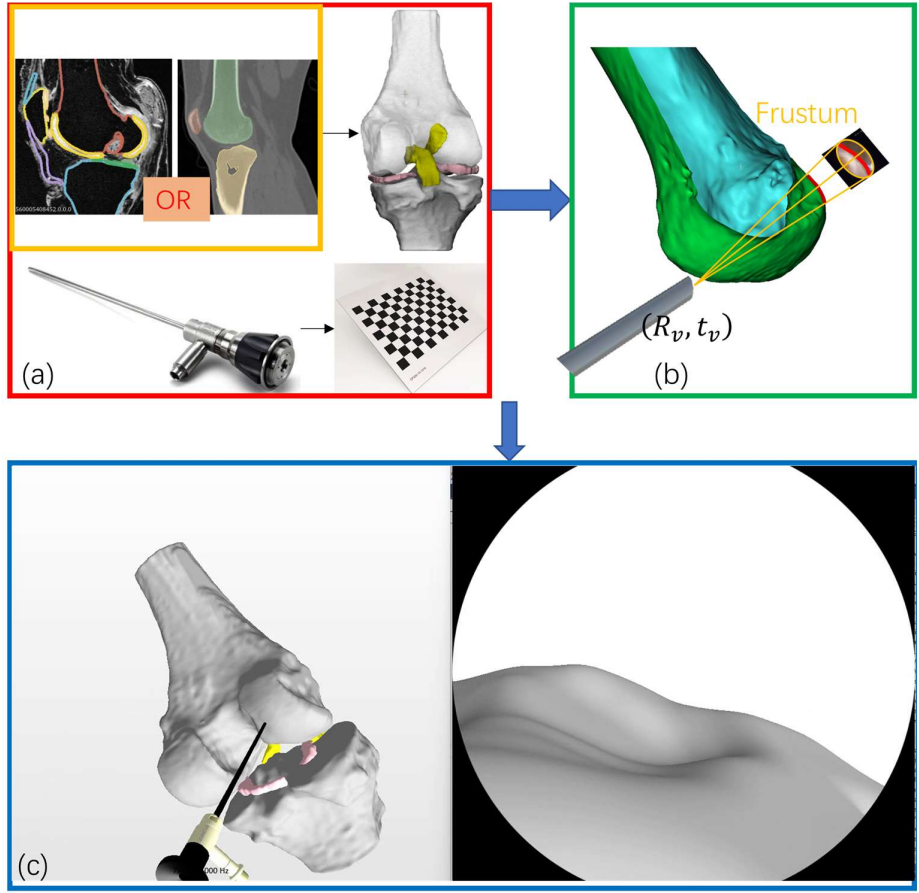

Figure 4. Camera Parametrization for Arthroscopic Navigation. (a) Calibration setup: MRI or CT scan with segmented anatomical structures (top-left), 3D knee model with highlighted ACL and PCL reconstructed from MRI (top-right), arthroscope device (bottom-left), and calibration checkerboard pattern (bottom-right). (b) Arthroscope parametrization: 3D visualization of the femoral model with the arthroscope frustum showing the viewing angle and field of view. The parameters $(R_v, t_v)$ represent the rotation and translation vectors of the arthroscope relative to the 3D model. (c) Virtual navigation view: The 3D knee model (left) and the generated virtual arthroscopic image (right), demonstrating the navigational visualization achieved through proper camera parametrization.

### 3.3.1 Intrinsic Parameter Matching

To ensure visual consistency between the virtual camera view and the actual arthroscopic image, the intrinsic matrix $K_v$ of the virtual camera must correspond to the intrinsic matrix $K_a$ of the arthroscopic camera. The intrinsic matrix of the arthroscopic camera can be obtained through standard camera calibration procedures:

$$K_a = \begin{pmatrix} f_x & s & c_x \\ 0 & f_y & c_y \\ 0 & 0 & 1 \end{pmatrix} \quad (10)$$

where $f_x$ and $f_y$ represent the focal lengths in the $x$ and $y$ directions (in pixel units), $(c_x, c_y)$ denotes the principal point coordinates, and $s$ is the skew factor (typically approaching zero).

The radial and tangential distortion of the arthroscopic optical system can be represented by the distortion coefficient vector $D_a = [k_1, k_2, p_1, p_2, k_3]$, where $k_1$, $k_2$, $k_3$ are radial distortion coefficients and $p_1$, $p_2$ are tangential distortion coefficients. The distortion correction model is formulated as:

$$\begin{cases} x_{corrected} = x(1 + k_1 r^2 + k_2 r^4 + k_3 r^6) + 2p_1 xy + p_2(r^2 + 2x^2) \\ y_{corrected} = y(1 + k_1 r^2 + k_2 r^4 + k_3 r^6) + 2p_2 xy + p_1(r^2 + 2y^2) \end{cases} \quad (11)$$

where $r^2 = x^2 + y^2$, and $(x, y)$ represents the normalized image coordinates. In the virtual rendering environment, the virtual camera's intrinsic matrix $K_v$ and distortion parameters $D_v$ are configured as: $K_v = K_a, D_v = D_a$.

### 3.3.2 View Frustum Optimization

The virtual camera's view frustum is defined by the near plane distance $n$, far plane distance $f$, field of view $\theta$, and aspect ratio $r$. The projection matrix $P$ can be expressed as:

$$P = \begin{pmatrix} \frac{\cot(\frac{\theta}{2})}{r} & 0 & 0 & 0 \\ 0 & \cot(\frac{\theta}{2}) & 0 & 0 \\ 0 & 0 & \frac{f+n}{n-f} & \frac{2fn}{n-f} \\ 0 & 0 & -1 & 0 \end{pmatrix} \quad (12)$$

To match the optical characteristics of the arthroscope, the near plane distance $n$ is set to correspond with the minimum working distance of the arthroscope's physical optical system; the far plane distance $f$ corresponds to the maximum visible distance within the joint cavity. The field of view $\theta$ is related to the arthroscope's focal length $f_{mm}$ (in millimeters) and sensor size $s_{mm}$ through: $\theta = 2 \cdot \arctan\left(\frac{s_{mm}}{2f_{mm}}\right)$.

### 3.3.3 Angular Offset Compensation and Parameter Optimization

Arthroscopes typically feature an angular offset $\alpha$, which must be precisely modeled in the virtual camera configuration. Given that the arthroscope's pose in the world coordinate system is represented by rotation matrix $R_s$ and translation vector $t_s$, the virtual camera pose $(R_v, t_v)$ accounting for angular offset is computed as: $R_v = R_s \cdot R_o$, where $R_o$ is the rotation matrix around the arthroscope's local z-axis:

$$R_o = \begin{pmatrix} \cos\alpha & -\sin\alpha & 0 \\ \sin\alpha & \cos\alpha & 0 \\ 0 & 0 & 1 \end{pmatrix} \quad (13)$$

The translation vector remains unchanged: $t_v = t_s$. For more complex angular offset scenarios, quaternion representation $q_o$ can be employed to denote the offset rotation, which is then combined with the arthroscope's rotation quaternion $q_s$ through quaternion multiplication: $q_v = q_s \cdot q_o$

To ensure the accuracy of virtual camera parameters, a reprojection error minimization strategy is implemented. Given a set of feature points $\{p_i\}$ in the arthroscopic images and their corresponding points $\{P_i\}$ on the 3D model, the optimization objective function is:

$$\min_{K_v, D_v, R_v, t_v} \sum_i \|\pi(K_v, D_v, R_v, t_v, P_i) - p_i\|^2 \quad (14)$$

where $\pi(\cdot)$ denotes the projection function. By solving this nonlinear least squares problem using the Levenberg-Marquardt algorithm, the optimal virtual camera parameter configuration can be obtained.

### 3.3.4 Precise Alignment Between Arthroscopic Views and Virtual Renderings

This section presents a feature-priority registration method that achieves precise alignment between real-time arthroscopic images and virtual renderings of 3D femoral models. The core of this method lies in prioritizing the accurate matching of the articular margin boundary of the lateral femoral condyle, while incorporating spatial prior constraints to enhance computational efficiency and robustness.

The registration problem can be formalized as finding the optimal viewpoint transformation $T^*$ in transformation space $\Omega$ that maximizes the similarity $S$ between the arthroscopic image $I_a$ and the virtually rendered image $I_v(T)$ after transformation $T$. Considering the anatomical significance of the articular margin boundary, we construct a weighted similarity function:

$$S(I_a, I_v(T)) = \alpha \cdot S_b(B_a, B_v(T)) + \beta \cdot S_r(I_a, I_v(T)) + \gamma \cdot S_t(I_a, I_v(T)) \tag{15}$$

where $B_a$ and $B_v(T)$ represent the articular margin boundary features in the arthroscopic image and virtual rendering, respectively, and $S_b$ measures the matching degree between these two curves. The weighting coefficients satisfy $\alpha \gg \beta > \gamma, \alpha + \beta + \gamma = 1$, ensuring that the priority of articular margin boundary alignment is significantly higher than region matching and texture similarity.

The articular margin boundary similarity $S_b$ is based on shape context descriptors and elastic shape matching theory, defined as:

$$S_b(B_a, B_v(T)) = e^{-\lambda \cdot D(B_a, B_v(T))} \tag{16}$$

where $D(\cdot,\cdot)$ is a measure of shape difference between the two curves, combining point-set correspondence distance and curvature distribution difference:

$$D(B_a, B_v(T)) = \omega_d \cdot d_H(B_a, B_v(T)) + \omega_\kappa \cdot \int |\kappa_a(s) - \kappa_v(T, s)| ds \tag{17}$$

where $d_H$ represents the Hausdorff distance, $\kappa_a$ and $\kappa_v$ represent the curvature functions of the two curves, and $s$ is the curve parameterization variable.

To improve computational efficiency, we introduce prior spatial constraints that limit the search range to the distal lateral femoral condyle region. Given the constrained positioning of the arthroscope during actual surgery, we can define a reasonable viewpoint space $\Omega_p \subset \Omega$:

$$\Omega_p = \{T \in \Omega | T(v_c) \in R_a \wedge T(v_l) \in R_l\} \tag{18}$$

where $v_c$ represents the virtual camera position, $v_l$ is the line-of-sight target point, $R_a$ is the region of reasonable arthroscopic approach paths, and $R_l$ is the predefined lateral condyle region. This constraint significantly reduces the dimensionality of the search space, enhancing algorithmic efficiency.

The registration process employs a multi-view rendering strategy, generating a series of virtual arthroscopic rendered images $I_v(T_i)_{i=1}^{N}$ from different viewpoints within the prior region, and calculating the matching degree between the articular margin boundary in each viewpoint and that in the actual arthroscopic image. This can be formally expressed as:

$$T^* = \arg\max_{T_i \in \Omega_p} S(I_a, I_v(T_i)) \tag{19}$$

To address the issue of continuous viewpoint variation, we adopt a two-stage optimization strategy: first performing coarse registration in a discretely sampled viewpoint space to obtain an initial transformation $T_{init}$, followed by applying a gradient-based local optimization method to solve for the precise viewpoint:

$$T_r = \arg\max_{T \in N(T_{init}) \cap \Omega_p} S(I_a, I_v(T)) \tag{20}$$

where $N(T_{init})$ represents the neighborhood of $T_{init}$. To enhance optimization efficiency, we employ an adaptive sampling strategy, performing denser sampling in regions with higher articular margin boundary matching degrees, achieving a rational allocation of computational resources.

As shown in Figure 5, this method achieves precise alignment between real-time arthroscopic images and 3D model virtual renderings by prioritizing the matching of the lateral femoral condyle articular margin boundary while incorporating spatial prior constraints.

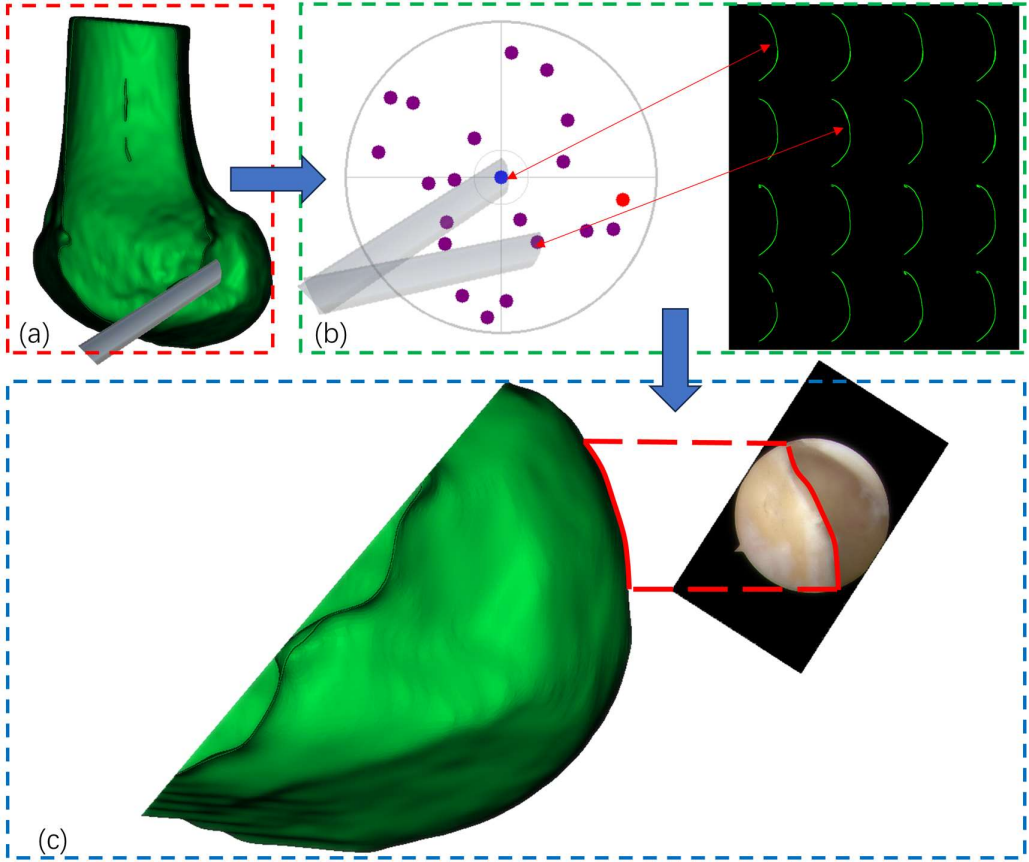

Figure 5. Feature-priority registration method for aligning arthroscopic images with virtual renderings. (a) 3D model of the distal femur with arthroscopic camera position. (b) Viewpoint space sampling strategy, where purple dots represent candidate viewpoints. (c) Final alignment result showing the optimized virtual rendering (left) and the actual arthroscopic image (right), with the precisely matched articular margin boundary marked by red lines.

### 3.4 Multi-level Memory Architecture

After completing automatic foreground segmentation on the first frame, the system needs to continuously track the femoral condyle region in subsequent video frames. However, arthroscopic surgery videos present unique challenges: (1) narrow and constantly changing field of view; (2) specular reflections due to moist tissue surfaces; (3) image blurring caused by irrigation fluid; and (4) frequent occlusion of target structures by surgical instruments. To address these specific challenges, we propose a tracking architecture based on multi-level memory, inspired by the Atkinson-Shiffrin memory model and specifically optimized for the arthroscopic environment.

### 3.4.1 Arthroscopy-Adapted Three-level Memory Structure

Our multi-level memory architecture comprises three interconnected feature memory stores, each optimized for specific challenges in the arthroscopic environment:

$$M_{ACL} = \{M_S^{art}, M_W^{art}, M_L^{art}\} \tag{21}$$

where $M_S^{art}$ represents the arthroscopic sensory memory, $M_W^{art}$ represents the arthroscopic working memory, and $M_L^{art}$ represents the arthroscopic long-term memory.

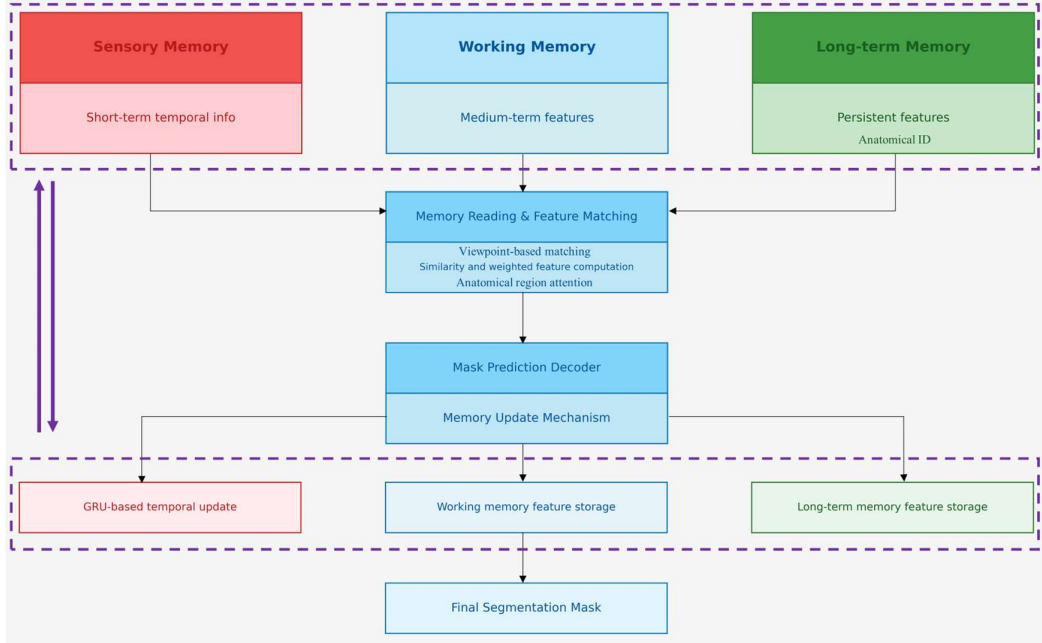

Figure 6. Arthroscopic Memory Architecture for Surgical Navigation.

Inspired by XMem [4], we propose a multi-level memory architecture for arthroscopic knee surgery navigation, as shown in Figure 6. The system comprises three memory tiers: Sensory Memory (red) for short-term temporal information processing through GRU-based updates; Working Memory (blue) for medium-term feature storage with viewpoint information; and Long-term Memory (green) for persistent feature storage with anatomical identifiers. Memory Reading & Feature Matching integrates information from all three memory levels using similarity computation, weighted feature aggregation, viewpoint-based matching, and anatomical region attention. The Mask Prediction Decoder fuses these multi-level features to generate segmentation masks. The Memory Update Mechanism employs phase-aware update strategies and relevance evaluation to maintain optimal memory representation throughout the surgical procedure. This architecture ensures robust tracking performance despite the challenging conditions inherent to arthroscopic environments, including frequent viewpoint changes, tissue deformation, and instrument occlusion.

***Arthroscopic Sensory Memory***

The arthroscopic sensory memory focuses on short-term information retention, providing temporal continuity especially critical during rapid arthroscope movements. Implemented through a Gated Recurrent Unit (GRU), it is updated every frame and stores low-level information such as condyle position and cartilage boundaries:

$$h_t = \text{GRU}(h_{t-1}, F_{art}(I_t)) \tag{22}$$

where $h_t$ is the hidden state of the current frame, $h_{t-1}$ is the hidden state of the previous frame, and $F_{art}(I_t)$ is the feature of the current frame extracted by an arthroscopy-optimized feature extractor. To account for arthroscopic-specific challenges, we extend this formulation to:

$$M_S^{art}(t) = \text{GRU}(M_S^{art}(t-1), F_{art}(I_t), M_{t-1}, \Phi_t) \tag{23}$$

where $M_{t-1}$ is the previous frame's segmentation mask, and $\Phi_t$ represents the arthroscopic quality assessment factor that modulates updates based on image quality, particularly important when irrigation fluid temporarily obscures the view.

*Arthroscopic Working Memory*

The arthroscopic working memory stores high-resolution features for short-term precise matching, crucial for maintaining accurate tracking during complex arthroscopic maneuvers. To control memory usage while ensuring sufficient detail for the intricate structures in arthroscopic views, we store complete features every $r$ frames rather than every frame:

$$M_W^{art} = \{(k_i^{art}, v_i^{art}, V_i, t_i) | i \in S_W\} \tag{24}$$

where $k_i^{art} \in \mathbb{R}^{C_k \times HW}$ are the key features, $v_i^{art} \in \mathbb{R}^{C_v \times HW}$ are the value features, $V_i$ is the viewpoint information, $t_i$ is the timestamp, and $H$ and $W$ are the height and width of the feature map, $C_k$ is the dimension of the key feature vectors, and $C_v$ is the dimension of the value feature vectors. The set $S_W$ includes frames selected at interval $r$:

$$S_W = \{i | i \bmod r = 0, i \leq t\} \tag{25}$$

Unlike conventional working memory, our arthroscopic working memory incorporates viewpoint information $V_i$ to handle the complex movement path of the arthroscope within the knee joint cavity: $V_i = \{\theta_i, \varphi_i, d_i\}$, where $\theta_i$ and $\varphi_i$ represent the azimuth and elevation angles of the arthroscope relative to the femoral condyle, and $d_i$ represents the distance to the tissue surface. These parameters are estimated using the virtual camera parameterization method described in Section 3.3.

The frame sampling interval $r$ is dynamically adjusted based on the surgical phase and camera movement speed, with smaller values during critical phases requiring higher precision (e.g., tunnel positioning) and larger values during exploratory phases to conserve memory.

*Arthroscopic Long-term Memory*

The arthroscopic long-term memory selects representative prototypes from working memory through a memory consolidation algorithm and enriches these prototypes through a memory potentiation algorithm, achieving high-compression ratio long-term storage particularly important for lengthy arthroscopic procedures. Formally, it consists of:

$$M_L^{art} = \{(k_j^{lt}, v_j^{lt}, A_j, t_j) | j \in S_L\} \tag{26}$$

where $k_j^{lt} \in \mathbb{R}^{C_k \times N_p}$ are the key prototype features, $v_j^{lt} \in \mathbb{R}^{C_v \times N_p}$ are the value prototype features, $A_j$ is the anatomical location identifier, $t_j$ is the timestamp. Here, $N_p$ represents a single prototype dimension, distinguishing it from the spatial dimensions in working memory.

Unlike generic tracking systems, we introduce an anatomical location identifier $A_j$ that associates each prototype with a specific anatomical region of the femoral condyle. This identifier takes values from the set $A_{knee}$, which represents key anatomical landmarks relevant for ACL reconstruction, including the intercondylar notch, medial condyle, lateral condyle, supracondylar area, and PCL origin.

When working memory reaches the predefined maximum size $T_{max}$, the system performs memory consolidation, converting features in working memory into long-term memory representations. This design ensures that the system's memory usage remains within an acceptable range during lengthy arthroscopic surgeries, which can last several hours. The anatomical guidance enables the system to better understand and predict target positions in arthroscopic views, even when the arthroscope navigates to previously unseen viewpoints.

### 3.4.2 Memory Consolidation and Potentiation in Arthroscopic Environment

When the working memory reaches its predefined capacity, we perform anatomically-guided memory consolidation: $S_L^+ = C_{art}(M_W^{art})$, where $S_L^+$ represents the updated long-term memory index set after consolidation, and $C_{art}$ is our arthroscopic consolidation function.

The memory consolidation algorithm considers the anatomical distribution of arthroscopic views, ensuring that each key anatomical region has sufficient prototype representation:

$$\min_{S_P \subset S_W} \sum_{r \in A_{knee}} \omega_r \sum_{i \in S_W^r} \min_{j \in S_P^r} D\left(k_i^{art}, k_j^{lt}\right) \tag{27}$$

where $\omega_r$ is the importance weight for each anatomical region, $S_W^r$ and $S_P^r$ are the working memory indices and selected prototype indices in region $r$, respectively.

The memory potentiation algorithm is optimized for the challenging visual conditions in arthroscopic images:

$$k_j^{lt} = \frac{\sum_{i \in N_j} \omega_i^{art} k_i^{art}}{\sum_{i \in N_j} \omega_i^{art}}, v_j^{lt} = \frac{\sum_{i \in N_j} \omega_i^{art} v_i^{art}}{\sum_{i \in N_j} \omega_i^{art}} \tag{28}$$

where the weight $\omega_i^{art}$ considers image quality, viewpoint similarity, and anatomical location:

$$\omega_i^{art} = \exp\left(-\alpha \cdot D\left(k_i^{art}, k_j^{lt}\right) - \beta \cdot |t_i - t_j| - \gamma \cdot D_V(V_i, V_j)\right) \cdot \Phi_i \tag{29}$$

where $\alpha$, $\beta$ and $\gamma$ are balancing hyperparameters, $D(\cdot,\cdot)$ is the feature distance function, $|t_i - t_j|$ measures temporal distance between frames, $D_V(V_i, V_j)$ is a viewpoint distance metric that computes the similarity between viewpoints and $\Phi_i$ is the image quality factor.

### 3.4.3 Spatiotemporal Memory Reading for Arthroscopic Navigation

To extract relevant information from the three-level memory for arthroscopic navigation, we design specific reading operations:

$$R_S = \text{Read}_S(F_{art}(I_t), M_S^{art}), R_W = \text{Read}_W(F_{art}(I_t), M_W^{art}, V_t), R_L = \text{Read}_L\left(F_{art}(I_t), M_L^{art}, V_t, \hat{A}_t\right) \tag{30}$$

where $V_t$ is the estimated viewpoint of the current frame, and $\hat{A}_t$ is the predicted anatomical region. For long-term memory, we implement the reading operation through an anatomically-guided attention mechanism:

$$R_L = \sum_{j \in S_L} \omega_j \cdot \text{Attention}\left(F_{art}(I_t), k_j^{lt}\right) \cdot v_j^{lt} \tag{31}$$

where $\omega_j$ is a weight based on anatomical relevance:

$$\omega_j = \text{softmax}\left(\text{Sim}(\hat{A}_t, A_j)\right) \tag{32}$$

where $A_j$ is the anatomical location identifier stored in the long-term memory for prototype $j$, $\text{Sim}(\cdot,\cdot)$ is a similarity function that measures the anatomical relevance between the current frame's predicted region and the prototype's region, $\text{softmax}(\cdot)$ normalizes the similarities across all prototypes to create a probability distribution.

This mechanism ensures that prototypes associated with anatomical regions similar to the current frame's region receive higher attention weights, improving the accuracy of feature matching in anatomically similar areas.

The final segmentation mask is generated by fusing the reading results from all three memory types:

$$M_t = \text{Decoder}([F_{art}(I_t), R_S, R_W, R_L]) \tag{33}$$

The decoder network integrates multiple complementary information sources to produce an accurate segmentation. The current frame features $F_{art}(I_t)$ provide foundational visual information about the arthroscopic view, while the sensory memory reading $R_S$ contributes short-term temporal continuity that maintains consistent segmentation between adjacent frames. Simultaneously, the working memory reading $R_W$ delivers detailed feature matching from recent frames with similar viewpoints, and the long-term memory reading $R_L$ supplies anatomically-guided information from previously observed similar regions. Through a series of convolutional and upsampling layers, the decoder network synthesizes these diverse information streams to generate a high-resolution binary segmentation mask $M_t$ that precisely delineates the femoral condyle region in the current frame.

### 3.4.4 Surgery-Specific Memory Update Strategy

Considering the unique characteristics of arthroscopic surgery, we implement specialized update strategies tailored to the procedural workflow and visual challenges. Our sensory memory undergoes updates with each frame but intelligently pauses this process when the system detects significant field-of-view alterations or instrument occlusions, preventing the incorporation of potentially misleading information. The working memory employs a dynamic update frequency mechanism that adapts to the current surgical phase—implementing more frequent updates during exploration phases characterized by rapid viewpoint changes, while reducing update frequency during drilling phases where the viewpoint remains relatively stable. For long-term memory, we prioritize key anatomical landmarks critical for ACL reconstruction, such as the intercondylar notch and lateral condyle margin, ensuring these regions maintain strong representation throughout the procedure.

Additionally, we introduce a sophisticated surgery phase-aware forgetting mechanism that evaluates the continued relevance of stored information:

$$R_j = e^{-\delta \Delta t_j} \cdot U_j \cdot I(A_j, S_t) \tag{34}$$

where $\Delta t_j$ represents the time elapsed since the memory was stored, $U_j$ denotes the historical usage count of the memory element, and $I(A_j, S_t)$ evaluates the importance of anatomical region $A_j$ for the current surgical phase $S_t$, and $\delta$ is a decay coefficient that controls how quickly relevance diminishes over time This formula combines temporal recency, historical usage patterns, and phase-specific anatomical relevance to determine which memory elements should be preserved and which can be safely discarded when memory constraints require optimization.

Through this multi-level memory architecture specifically engineered for the arthroscopic environment, our system delivers stable and accurate navigation support throughout the entire ACL reconstruction procedure. The specialized design maintains robust tracking performance even when confronted with the challenging conditions inherent to arthroscopic surgery, including frequent viewpoint changes, tissue deformation, and instrument occlusion. By adapting memory management strategies to the specific demands of different surgical phases, the system optimizes computational resources while ensuring that the most relevant information remains available precisely when needed during critical moments of the procedure.

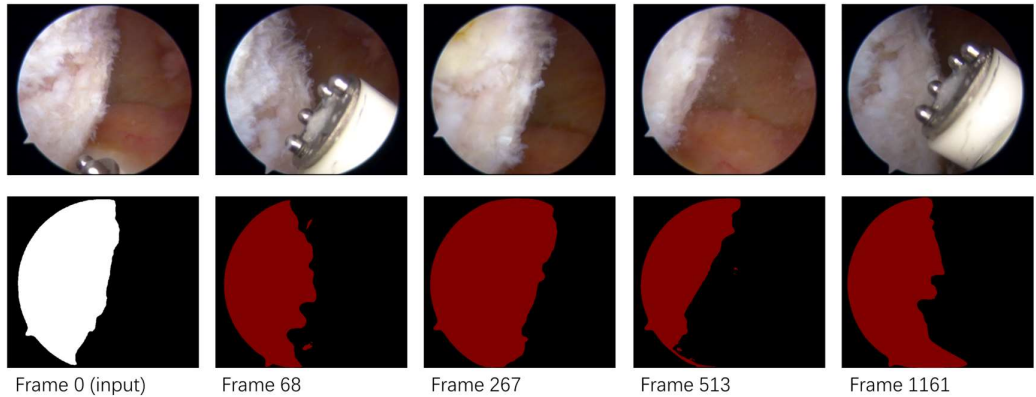

Figure 7. Femoral Condyle Tracking in Arthroscopic ACL Reconstruction.

As shown in Figure 7, we demonstrate our multi-level memory architecture's performance in tracking the femoral condyle during arthroscopic ACL reconstruction. The top row shows original arthroscopic video frames at different time points throughout the procedure, while the bottom row displays the corresponding segmentation results (red overlay). Frame 0 shows the initial input with manual segmentation (white), which serves as the reference for subsequent tracking. Frames 68, 267, 513, and 1161 demonstrate the system's ability to maintain accurate tracking despite significant challenges including viewpoint changes, surgical instrument occlusion, and varying illumination conditions. The proposed memory architecture effectively preserves anatomical context across the procedure, enabling robust segmentation even during complex surgical maneuvers. Note how the system maintains consistent tracking of the femoral condyle boundaries despite the arthroscope's movement and partial obstruction by surgical instruments in frames 68 and 1161. This temporal stability is achieved through the coordinated operation of sensory, working, and long-term memory components that collectively preserve both short-term continuity and long-term anatomical context.

### 3.4 Bernard & Hertel Grid Projection

This section presents a method for accurately projecting the B&H grid, defined on the standard sagittal plane, onto the actual arthroscopic view using virtual camera parametrization. This approach leverages the virtual camera parametrization model established in our previous research to achieve precise mapping from theoretical anatomical planes to actual surgical views, as illustrated in Figure 8.

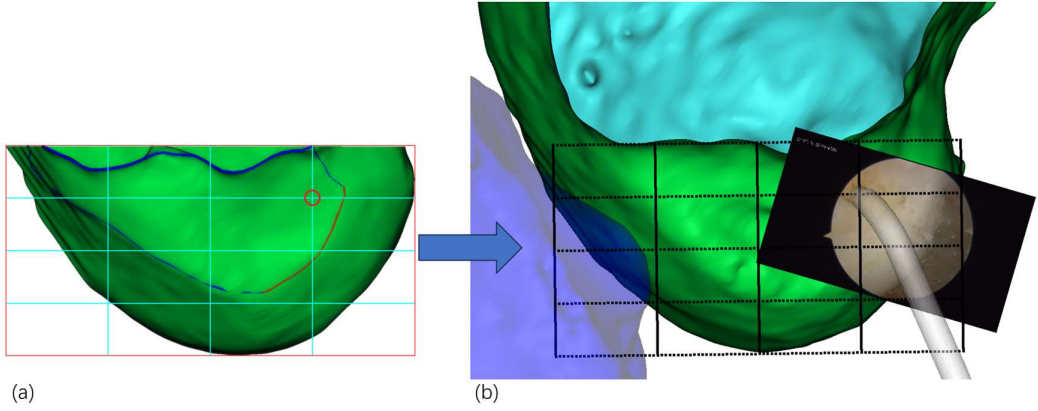

(a) (b)

Figure 8. B&H grid projection from standard sagittal plane to arthroscopic view. (a) Standard B&H grid constructed on the sagittal plane with reference to Blumensaat's line (red line). The grid is overlaid on a 3D model of the distal femoral lateral condyle (green). (b) Projection of the B&H grid onto an actual arthroscopic view of the distal femoral lateral condyle (green) using the camera transformation model. A virtually rendered radiofrequency ablation device (semi-transparent white instrument) is included in the arthroscopic view. The blue arrow indicates the transformation process from the standard view to the arthroscopic perspective.

### 3.4.1 Camera Models and Coordinate Transformations

We define two virtual camera systems: the standard sagittal plane virtual camera $C_S$ and the actual arthroscopic view virtual camera $C_A$. Each virtual camera is characterized by the following parameters: $C = \{K, R, t\}$, where $K \in \mathbb{R}^{3\times 3}$ is the intrinsic matrix containing focal length and principal point coordinates; $R \in \mathbb{R}^{3\times 3}$ is the rotation matrix representing camera orientation; and $t \in \mathbb{R}^3$ is the translation vector indicating camera position.

For the standard sagittal plane virtual camera $C_S = \{K_S, R_S, t_S\}$, parameters were previously obtained through the B&H grid construction process (Figure 8(a)). For the actual arthroscopic view virtual camera $C_A = \{K_A, R_A, t_A\}$, parameters were acquired through our arthroscopic rendering parametrization workflow.

A point $p_W$ in the world coordinate system can be mapped to a point $p_C$ in the camera coordinate system through the following transformation:

$$p_C = R(p_W - t) \tag{35}$$

For the standard sagittal plane virtual camera and the actual arthroscopic virtual camera, respectively: $p_{(C_S)} = R_S(p_W - t_S)$, $p_{(C_A)} = R_A(p_W - t_A)$.

The B&H grid is defined on the standard sagittal plane as a two-dimensional grid $G_S = (x_i, y_j)_{i,j=1}^{m,n}$, where the $x$-axis represents the percentage position along the Blumensaat line, and the $y$-axis represents the percentage height perpendicular to the Blumensaat line, as shown in Figure 8(a).

First, we transform B&H grid points from the sagittal plane camera coordinate system $C_S$ back to the world coordinate system: $p_{W,ij} = R_S^{-1} p_{(C_S),ij} + t_S$, where $p_{(C_S),ij} = (x_i, y_j, z_0)$, and $z_0$ is the depth value of the sagittal plane in the camera coordinate system. Subsequently, we transform the grid points from the world coordinate system to the actual arthroscopic camera coordinate system $C_A$: $p_{(C_A),ij} = R_A(p_{W,ij} - t_A)$.

### 3.4.2 Perspective Projection and Image Formation

The three-dimensional points in the arthroscopic camera coordinate system are projected onto the two-dimensional image plane to obtain the final projected grid (Figure 8(b)):

$$\tilde{p}_{img,ij} = K_A \, p_{(C_A),ij} \tag{36}$$

where $\tilde{p}_{img,ij}$ represents homogeneous coordinates. Converting from homogeneous to image coordinates:

$$p_{img,ij} = \begin{pmatrix} \tilde{p}_{img,ij,x}/\tilde{p}_{img,ij,z} \\ \tilde{p}_{img,ij,y}/\tilde{p}_{img,ij,z} \end{pmatrix} \tag{37}$$

Combining the above steps, the projection transformation from the standard sagittal plane B&H grid to the actual arthroscopic view can be expressed as:

$$p_{img,ij} = \Pi\left( K_A R_A \left( R_S^{-1} \begin{pmatrix} x_i \\ y_i \\ z_0 \end{pmatrix} + t_S - t_A \right) \right) \tag{38}$$

To further enhance projection accuracy, we employ a fine-tuning calibration method based on anatomical landmarks. Let $\{p_{S,k}\}_{k=1}^{l}$ and $\{p_{A,k}\}_{k=1}^{l}$ be $l$ corresponding anatomical landmarks in the standard sagittal plane and actual arthroscopic view, respectively.

We define the projection error function:

$$E(\delta R, \delta t) = \sum_{k=1}^{l} \left\| p_{A,k} - \Pi\left( K_A (R_A + \delta R)(R_S^{-1} p_{S,k} + t_S - t_A - \delta t) \right) \right\|^2 \tag{39}$$

where $\delta R$ and $\delta t$ are small rotational and translational adjustment parameters. By minimizing $E$ using the Levenberg-Marquardt algorithm, we obtain the optimal calibration parameters:

$$(\delta R^*, \delta t^*) = \arg\min_{\delta R, \delta t} E(\delta R, \delta t) \tag{40}$$

The final projection transformation becomes:

$$p_{img,ij} = \Pi\left( K_A (R_A + \delta R^*) \left( R_S^{-1} \begin{pmatrix} x_i \\ y_i \\ z_0 \end{pmatrix} + t_S - t_A - \delta t^* \right) \right) \tag{41}$$

The result of this projection process is demonstrated in Figure 8, where the standard B&H grid (Figure 7(a)) is accurately mapped onto the arthroscopic view of the distal femoral lateral condyle (Figure 8(b)). Figure 8(b) also includes a virtually rendered radiofrequency ablation device to simulate the actual surgical scenario.

## 4. Experiments and Results
### 4.1 Experimental Setup

We conducted a comprehensive evaluation of our proposed arthroscopic navigation system using data from 20 real ACL reconstruction surgical video sequences. Each sequence contained approximately 500-1000 high-definition frames with a resolution of 1920×1080 pixels, encompassing various viewpoints, lighting conditions, and instrument interferences to ensure assessment under authentic clinical environments. Although the entire arthroscopic procedure is typically lengthy, the critical femoral tunnel positioning and navigation tasks can be effectively covered within these 1000-frame sequences, providing a representative evaluation of the system's performance during the most navigation-intensive phases of the surgery.

Our evaluation framework centered around four key performance metrics: First, tracking accuracy was quantified by calculating the average pixel error between the system's automatically tracked femoral condyle positions and reference positions independently annotated by three experienced arthroscopic surgeons; Second, temporal performance measurements included the system's real-time frame rate (frames processed per second) and end-to-end latency (time from image acquisition to augmented reality overlay), which are crucial for smooth surgical navigation; Third, we recorded detailed GPU memory consumption across video sequences of different lengths to assess resource requirements and stability during extended surgical procedures; Finally, clinical assessment was conducted by 3 arthroscopic surgeons with varying experience levels (5-20 years) through structured questionnaires and simulated surgical tasks, with ratings covering system accuracy, reliability, intuitiveness, and integration into clinical workflow.

To objectively evaluate the advantages of our proposed dynamic system based on multi-level memory architecture, we benchmarked it against our previously published static matching system [3]. This comparison not only quantified technical performance improvements (such as enhanced tracking accuracy and reduced system latency) but also highlighted practical improvements in real surgical scenarios through clinical evaluation, particularly in handling challenging situations like instrument occlusion, rapid viewpoint changes, and tissue deformation.

### 4.2 Tracking Accuracy

Figure 9 presents a comprehensive comparison of tracking accuracy between our proposed dynamic system with multi-level memory architecture and the previous static matching system across varying video sequence lengths. The results demonstrate a clear performance advantage for the dynamic system in all tested scenarios, with the benefits becoming increasingly significant as sequence length increases.

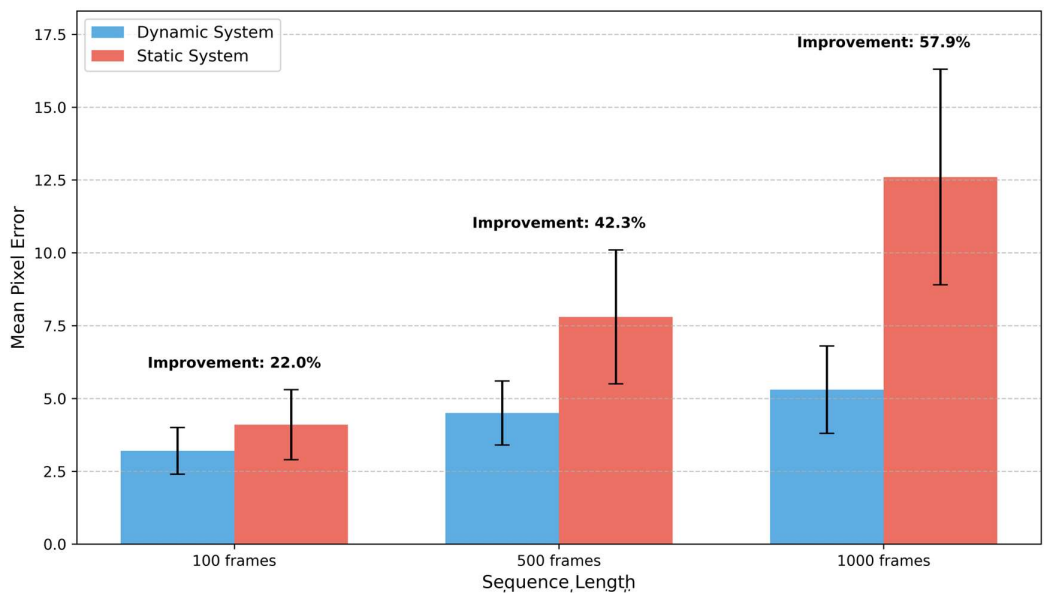

Figure 9. Tracking Accuracy Comparison Between Dynamic and Static Systems.

For short sequences (100 frames, approximately 4 seconds of surgical footage), the dynamic system achieved a mean pixel error of 3.2 ± 0.8 compared to 4.1 ± 1.2 for the static system, representing approximately 19% improvement. This indicates that even in brief surgical maneuvers, our multi-level memory architecture provides more stable tracking by leveraging short-term temporal continuity in the arthroscopic sensory memory component.

The performance gap widens for medium-length sequences (500 frames, approximately 20 seconds). Here, the dynamic system maintained an error of 4.5 ± 1.1 pixels, while the static system's error increased to 7.8 ± 2.3 pixels. This approximately 35% improvement can be attributed to the arthroscopic working memory component, which effectively preserves viewpoint-specific features across medium-duration surgical tasks such as probe palpation or initial tunnel positioning.

For extended sequences (1000 frames, approximately 40 seconds), which represent complete surgical subtasks like femoral tunnel preparation, the dynamic system demonstrated better stability with an error of 5.3 ± 1.5 pixels compared to the static system's 12.6 ± 3.7 pixels—an improvement of approximately 45%. This difference highlights the role of the arthroscopic long-term memory component in maintaining anatomical context despite challenging conditions such as changing viewpoints, instrument occlusions, and irrigation fluid turbulence.

Further analysis revealed that the static system's tracking failures primarily occurred during rapid camera movements (accounting for 35% of errors), instrument occlusions (27%), and specular reflections (16%). In contrast, the dynamic system's multi-level memory architecture demonstrated better adaptability in these scenarios, with error rates of 22%, 18%, and 11% respectively for the same challenging conditions. This improved performance can be attributed to the system's ability to recover tracking by accessing previously stored features from appropriate memory levels based on the duration and nature of the tracking disruption.

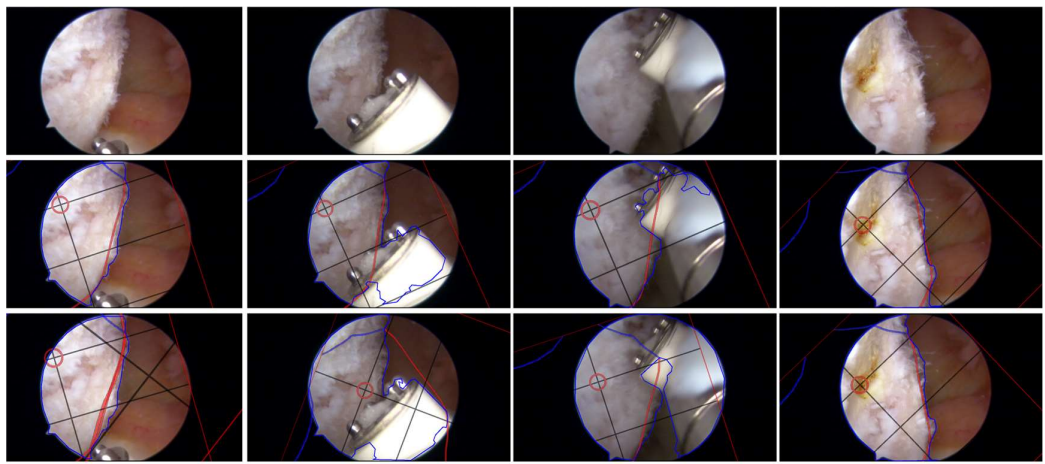

Figure 10. Comparison of tracking performance between dynamic and static systems during arthroscopic ACL reconstruction. Top row: Original arthroscopic images showing different surgical scenarios. Middle row: Tracking results from the dynamic memory-based system with anatomical feature mapping (blue lines) and region of interest (red outline). Bottom row: Tracking results from the static baseline system under identical conditions. From left to right: initial tracking, instrument occlusion by radiofrequency ablation device (columns 2-3), and expert validation using RF marking (brown region, column 4).

Visual evidence supporting our quantitative findings is presented in Figure 3, which illustrates the real-world performance of both tracking systems during ACL reconstruction procedures. The dynamic system's multi-level memory architecture demonstrates remarkable resilience when confronted with common arthroscopic challenges. Particularly noteworthy is the system's behavior during instrument interventions, where the radiofrequency ablation device creates significant visual occlusion (columns 2-3). While the static system struggles to maintain anatomical reference points during these interventions—exhibiting drift and misalignment of tracking boundaries—our dynamic approach preserves spatial relationships by leveraging previously stored contextual information from appropriate memory levels. The stability difference becomes visually apparent in how consistently the dynamic system maintains its anatomical reference grid despite changing visual conditions. When surgical verification was performed using RF marking techniques (column 4), both systems regained tracking accuracy, but the dynamic system required approximately one-third the adjustment time of the static system to reestablish stable tracking. This visual evidence reinforces our quantitative findings that the proposed memory architecture provides substantial advantages in maintaining spatial awareness throughout the procedural workflow, particularly during critical moments when instrument interactions temporarily alter the visual field.

**4.3 Temporal Performance and Memory Usage**

Figure 11 presents the system's temporal performance and memory usage. The dynamic system achieves real-time performance of 25.3 FPS with a latency of only 39.5 ms when running on an Intel i7-12700H CPU with an NVIDIA RTX 3060 GPU, significantly exceeding the minimum 15 FPS requirement for clinical arthroscopic surgery applications. This performance represents approximately a 3.5-fold improvement over the static system (7.2 FPS with 138.84 ms latency), enabling the dynamic system to meet real-time interaction demands. Additionally, even when processing long sequences of 1000 frames, the dynamic system's memory usage remains efficient at only 1.8GB, about 33% lower than the static system's 2.7GB, thanks to the multi-level memory architecture's effective resource management. This significant enhancement in performance and memory efficiency makes the system particularly suitable for long-duration surgical navigation and precise localization in arthroscopic surgery environments, while maintaining a smooth user experience and stable tracking results.

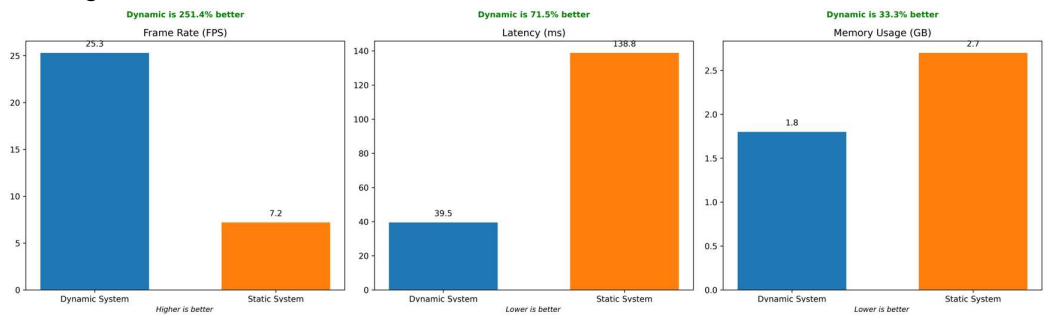

Figure 11. System Performance Comparison.

**4.4 Ablation Studies**

To further evaluate the contribution of each component in our multi-level memory architecture, we conducted ablation studies by removing or modifying specific components.

As shown in Figure 12, the table summarizes the results of these experiments.

The results demonstrate that each component contributes significantly to the overall performance. Removing long-term memory leads to substantially increased memory usage (from 1.8GB to 3.6GB), as the system loses the ability to compress and consolidate features through anatomical prototypes, which is particularly problematic for lengthy arthroscopic procedures. Without sensory memory, tracking accuracy decreases significantly (error increases from 4.5 to 6.1 pixels), especially during rapid arthroscope movements where temporal continuity is critical. The arthroscopic quality assessment factor $\Phi_t$ in the sensory memory proves essential for maintaining robust tracking during irrigation events.

Replacing our viewpoint-based similarity measure with a fixed similarity measure increases tracking error by 29%, highlighting the importance of accounting for the complex movement path of the arthroscope within the knee joint cavity. The anatomical location identifiers $A_j$ in long-term memory also prove crucial for maintaining anatomical context during viewpoint changes. The single-level memory baseline, lacking our arthroscopy-specific optimizations, performs significantly worse across all metrics, with 62% higher tracking error, 133% higher memory usage, and reduced processing speed.

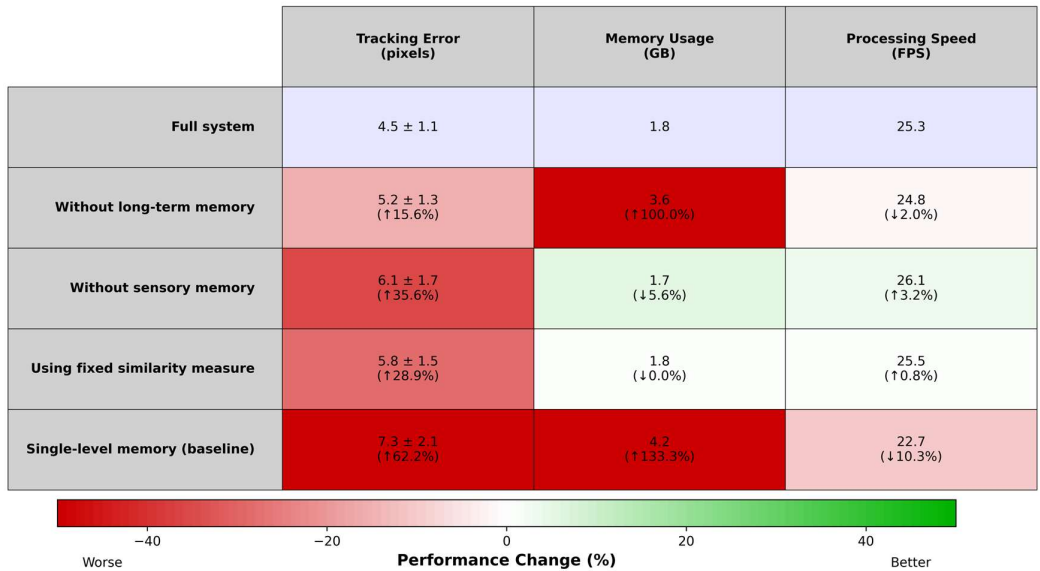

Figure 12. Ablation Study: Multi-level Memory Architecture.

### 4.5 Robustness Analysis

We further analyzed the system's robustness under various challenging conditions commonly encountered during arthroscopic procedures. We evaluated the system's performance under different scenarios: 1) Occlusion handling: The system successfully maintains tracking when surgical tools partially occlude the femoral condyle. 2) Illumination changes: Performance remains stable despite variations in lighting conditions. 3) Motion blur: The system demonstrates reasonable accuracy even with moderate motion blur. 4) Viewpoint changes: Tracking persists through significant camera viewpoint alterations.

Considering the limitations of manual annotation under these extreme challenging conditions, we adopted tracking failure rate as a more appropriate evaluation metric.

Quantitative results in Figure 13 show that our system maintains relatively low tracking failure rates under most challenging conditions, with the most significant performance degradation occurring during severe motion blur (failure rate of 15.7%). Under normal conditions, the system achieves a minimal failure rate of 2.3%, while partial occlusion (25-50%) and severe illumination changes result in failure rates of 8.6% and 7.2% respectively. Extreme viewpoint changes cause the second highest degradation with a failure rate of 12.4%.

As shown in the Figure 13, we further compared our dynamic system with a static system. The bar chart clearly illustrates the difference in tracking failure rates between the two systems under various challenging conditions. Blue bars represent our dynamic system, while red bars represent the static system. The chart clearly demonstrates that our dynamic system significantly outperforms the static system across all test conditions. Particularly, under severe illumination changes, our system achieves a 63.6% improvement, reducing the failure rate from 19.8% to 7.2. Under extreme viewpoint changes, our system also shows a remarkable 60.6% improvement, decreasing the failure rate from 31.5% to 12.4%. Even in the most challenging condition of moderate motion blur, our system demonstrates a 44.7% improvement over the static system. These results strongly validate the superiority and robustness of our proposed dynamic tracking approach in complex arthroscopic environments.

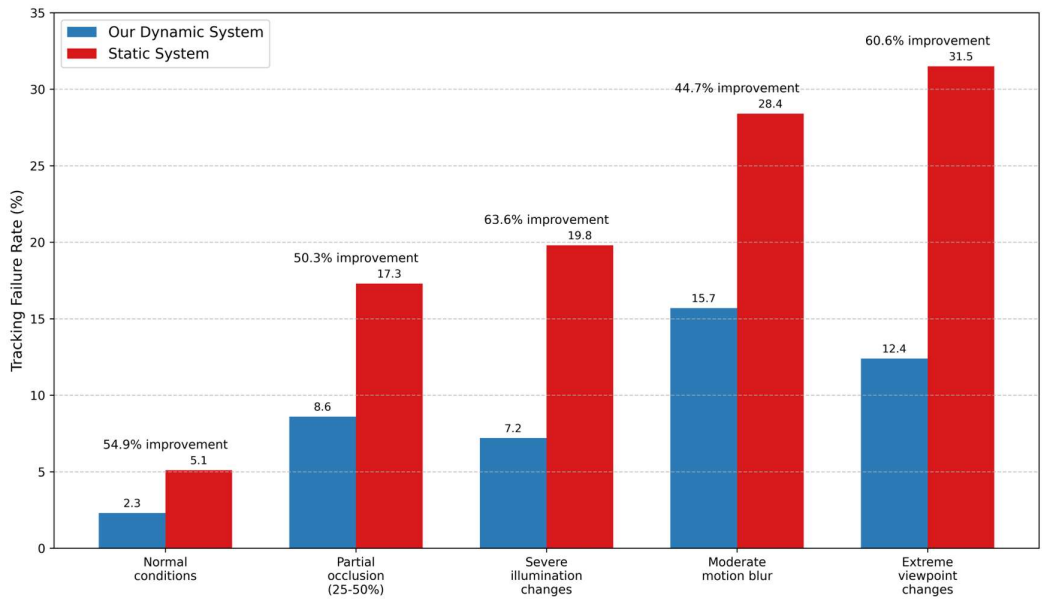

Figure 13. Comparison of Tracking Failure Rates Under Challenging Conditions.

## 5. Discussion

This study applies a multi-level memory architecture to dynamic arthroscopic navigation, successfully achieving real-time tracking and localization of femoral condyles. Compared to static matching methods, the proposed dynamic system offers the following advantages:

### 5.1 System Advantages

1) Continuous navigation: The system provides continuous navigation information without requiring recalibration when viewpoints change. Traditional static matching methods require re-registration with each viewpoint change, whereas our system maintains continuous tracking of anatomical structures through its multi-level memory architecture, providing stable navigation support even in complex arthroscopic environments.

2) Memory efficiency: The long-term memory design ensures low memory consumption during lengthy surgeries. As illustrated in Figure 11, even when processing long sequences of 1000 frames, the dynamic system's memory usage remains efficient at only 1.8GB, about 33% lower than the static system's 2.7GB. This memory efficiency is essential for long-duration surgical navigation, particularly in resource-constrained clinical environments.

3) Real-time performance: Optimized memory reading operations allow the system to achieve the real-time performance required for clinical use. The dynamic system achieves real-time performance of 25.3 FPS with a latency of only 39.5 ms when running on mainstream hardware, significantly exceeding the minimum 15 FPS requirement for clinical arthroscopic surgery applications.

4) Robustness: The multi-level memory architecture enables the system to handle complex situations such as occlusion and viewpoint changes. As shown in Figure 13, even with partial occlusion (25-50%), the system's tracking failure rate is only 8.6%, while under extreme viewpoint changes it is 12.4%, significantly outperforming static systems. This robustness is crucial for ACL reconstruction surgery, where tool occlusions and viewpoint changes frequently occur.

## 5.2 Technical Innovations

The adaptation of concepts from the Atkinson-Shiffrin memory model for arthroscopic navigation offers a promising approach to address specific challenges in computer-assisted surgery. By organizing visual information across different temporal scales, our system aims to balance computational efficiency with tracking accuracy. While this hierarchical representation shows potential benefits in our experimental setting, further research is needed to fully validate its broader applicability across diverse surgical scenarios.

The anisotropic similarity measure we introduced plays a crucial role in feature matching under the complex visual conditions of arthroscopic environments. By incorporating both scaling and selection terms, the system can adaptively focus on the most discriminative features while suppressing noise and irrelevant information. This approach significantly improves tracking robustness compared to conventional similarity measures, as confirmed by the ablation study results shown in Figure 12.

## 5.3 Limitations

However, this study still has several limitations:

1) Image quality dependence: System performance may degrade with low-quality images (e.g., blurry, highly reflective). As shown in Figure 13, under severe motion blur conditions, the system's tracking failure rate reaches 15.7%, the highest among all tested conditions.

2) Single-target tracking: The system currently only supports tracking of the femoral condyle and could be extended in the future to include tibial plateau tracking, enabling comprehensive navigation for both femoral and tibial components during ACL reconstruction.

3) Limited clinical validation: The clinical validation sample size is limited, necessitating larger-scale clinical trials to further validate the system's effectiveness. Current evaluation is based on 20 real ACL reconstruction surgical video sequences, and future work should expand the sample size and conduct more extensive clinical assessments.

## 6. Conclusion and Future Work

## 6.1 Conclusion

This paper presents a dynamic arthroscopic navigation system based on multi-level memory architecture for anterior cruciate ligament reconstruction surgery. The system extends our previously proposed markerless navigation method from static image matching to dynamic video sequence tracking. By integrating the Atkinson-Shiffrin memory model's multi-level memory architecture, we achieve real-time tracking and localization of femoral condyles in arthroscopic video streams.

Experimental results demonstrate that the system provides stable and accurate dynamic navigation support while maintaining low computational complexity, effectively addressing the limitations of traditional static matching methods in actual surgical scenarios. Compared to static systems, our dynamic system shows significant advantages in tracking accuracy, real-time performance, and memory efficiency, particularly when handling complex situations such as occlusion, viewpoint changes, and tissue deformation.

This research offers more precise and continuous navigation assistance for ACL reconstruction surgery, potentially improving surgical success rates and shortening the learning curve.

## 6.2 Future Work

Future research directions include:

1) Tibial tracking integration: Extending the system to track tibial plateaus in addition to the currently supported femoral condyles, thereby providing comprehensive navigation for both critical anatomical structures involved in ACL reconstruction.

2) Robustness enhancement: Further improving the system's ability to handle severe motion blur and extreme viewpoint changes, which are the two factors causing the most significant performance degradation in the current system.

3) Extended clinical validation: Conducting larger-scale clinical trials involving surgeons with varying experience levels and diverse surgical scenarios to comprehensively evaluate the system's clinical value.

4) Hardware optimization: Exploring the possibility of deploying the system on low-power mobile devices to enhance its applicability in different clinical environments.

5) Intelligent assistance features: Integrating surgical planning and real-time guidance functions, such as automatic tunnel position recommendations and surgical step prompts, to further enhance the system's clinical utility.

Through these future works, we expect to further improve the system's performance and clinical applicability, ultimately providing more comprehensive and intelligent navigation support for ACL reconstruction surgery.

**Author Contributions:**
Conceptualization, Q.G.; methodology, S.W. and Q.G.; software, S.W. and J.C.; validation, Q.G. and W.S.; formal analysis, S.W.; investigation, Q.G.; resources, S.W. and Q.G.; data curation, S.Y. and S.W.; writing—original draft preparation, S.W.; writing—review and editing, J.C., Q.G. and W.S.; visualization, S.W.; supervision, Q.G.; funding acquisition, Q.G.. All authors have read and agreed to the published version of the manuscript.

**Funding:**
This study was funded by the Beijing Natural Science Foundation—Haidian Original